\documentclass[conference]{IEEEtran}
\IEEEoverridecommandlockouts
\usepackage{cite}
\usepackage{amsmath,amssymb,amsfonts}
\usepackage{algorithmic}
\usepackage{graphicx}
\usepackage{textcomp}
\usepackage{xcolor}
\usepackage{booktabs}
\def\BibTeX{{\rm B\kern-.05em{\sc i\kern-.025em b}\kern-.08em
    T\kern-.1667em\lower.7ex\hbox{E}\kern-.125emX}}
\begin{document}


\title{Remaining Useful Life Estimation of Hard Disk Drives using Bidirectional LSTM Networks\\}



\author{\IEEEauthorblockN{Austin Coursey\IEEEauthorrefmark{1},
Gopal Nath\IEEEauthorrefmark{2}, Srikanth Prabhu\IEEEauthorrefmark{3} and
Saptarshi Sengupta\IEEEauthorrefmark{1}}
\IEEEauthorblockA{\IEEEauthorrefmark{1}Department of Computer Science and Information Systems, Murray State University, Murray, KY, USA \\
\IEEEauthorrefmark{2}Department of Mathematics and Statistics, Murray State University, Murray, KY, USA\\
\IEEEauthorrefmark{3}Department of Computer Science and Engineering, Manipal Institute of Technology, Manipal, KA, India\\
Email: \IEEEauthorrefmark{1}acoursey3@murraystate.edu,
\IEEEauthorrefmark{2}gnath@murraystate.edu,
\IEEEauthorrefmark{3}srikanth.prabhu@manipal.edu,
\IEEEauthorrefmark{1}ssengupta@murraystate.edu}}


\maketitle

\begin{abstract}

Physical and cloud storage services are well-served by functioning and reliable high-volume storage systems. Recent observations point to hard disk reliability as one of the most pressing reliability issues in data centers containing massive volumes of storage devices such as HDDs. In this regard, early detection of impending failure at the disk level aids in reducing system downtime and reduces operational loss making proactive health monitoring a priority for AIOps in such settings. In this work, we introduce methods of extracting meaningful attributes associated with operational failure and of pre-processing the highly imbalanced health statistics data for subsequent prediction tasks using data-driven approaches. We use a Bidirectional LSTM with a multi-day look back period to learn the temporal progression of health indicators and baseline them against vanilla LSTM and Random Forest models to come up with several key metrics that establish the usefulness of and superiority of our model under some tightly defined operational constraints. For example, using a 15 day look back period, our approach can predict the occurrence of disk failure with an accuracy of 96.4\% considering test data 60 days before failure. This helps to alert operations maintenance well in-advance about potential mitigation needs. In addition, our model reports a mean absolute error of 0.12 for predicting failure up to 60 days in advance, placing it among the state-of-the-art in recent literature.


\end{abstract}


\section{Introduction}
Hard disk drives or HDDs have been the primary storage systems in large-scale data centers. Like any electronic device, HDDs also have a limited lifespan. To monitor operational health of these devices, operators frequently rely on S.M.A.R.T. (Self-Monitoring Analysis and Reporting Technology). S.M.A.R.T. logs device health data such as power-on hours, temperature and sector error rates, head flying hours and so on. \cite{smart}. These are set by each drive manufacturer along with a threshold for each attribute which is traditionally used to detect failure. If a hard drive is operating as it should, each S.M.A.R.T. attribute should ideally not be outside this threshold.

Backblaze, a cloud storage and data backup company, has over 171,000 hard drives in their data center \cite{backblaze}. With an annualized failure rate of 0.93\% in 2020, over a thousand of their hard drives failed \cite{klein_2021}. It follows that it is necessary for hard drive manufacturers and companies like Backblaze to know the remaining useful life (RUL) of their hard drives. This would not only help reduce downtime at a large data center, but would help protect valuable user data.

S.M.A.R.T. attributes are a great way of detecting imminent failure of hard drives \cite{beach_2021}, but it would be ideal to predict the RUL of a hard drive. This would allow consumers to know that they need to start making backups and initiate planning to buy a new hard drive long before it fails. It would also allow companies like Backblaze to prepare to replace a drive ahead of time. 

\begin{figure*}[ht!]
\begin{center}
    \makebox[\linewidth]{
\includegraphics[scale=0.5]{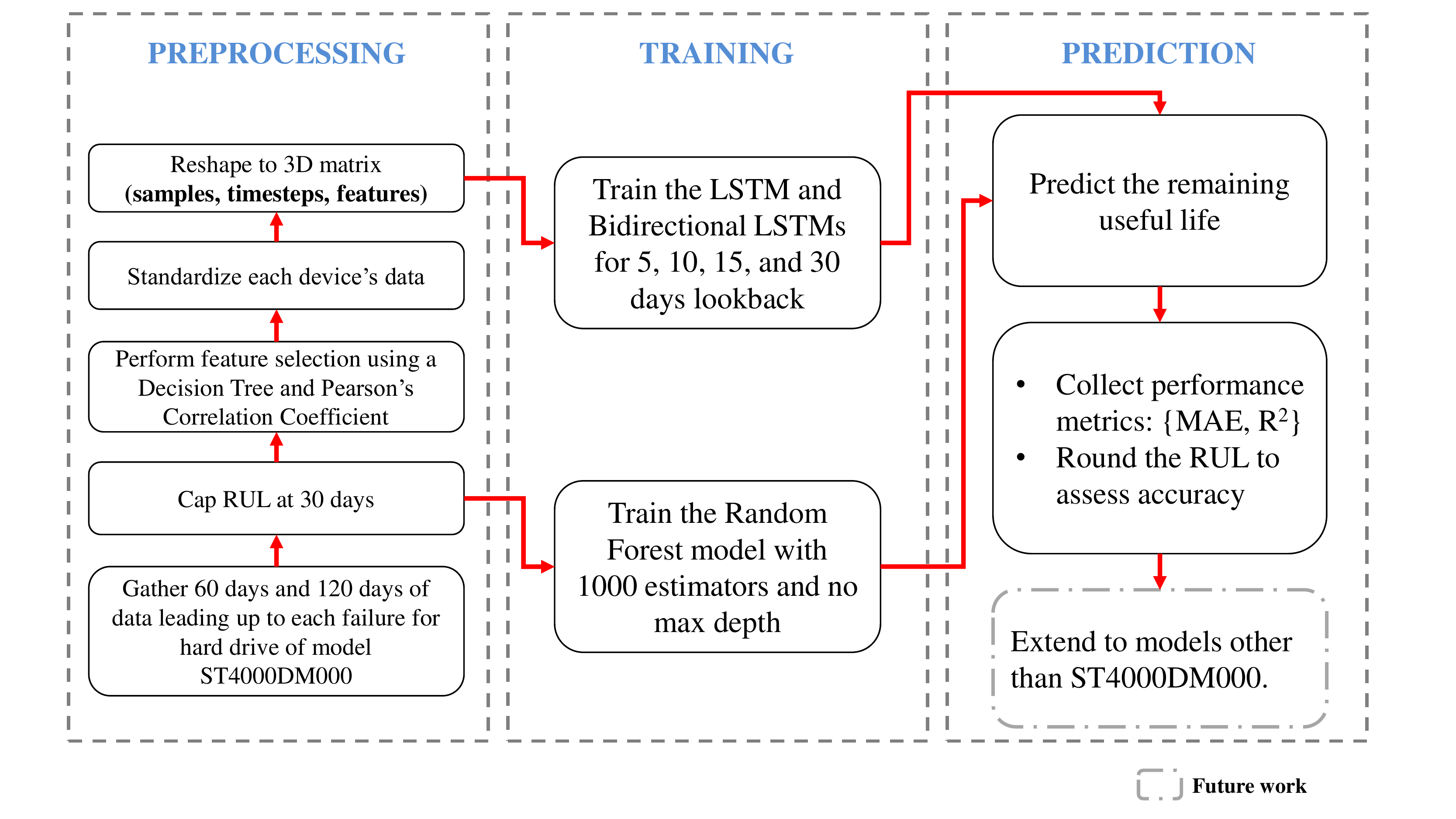}
}
\caption{Overview of the workflow for our approach.}

\label{fig:workflow}

\end{center}
\end{figure*}

\textbf{Contributions:} In this work, we propose methods for data standardization, normalization and RUL prediction for working with the highly class-imbalanced Backblaze data using a Bidirectional Long Short Term Memory Network \cite{bidirection_lstm} with multiple days of look-back period. Our approach considers S.M.A.R.T. features which are highly correlated to failure and builds a prediction pipeline that takes into consideration the long-term temporal relations in the failure data. We employ a vanilla Long Short Term Memory Network under similar data preprocessing conditions to contend that the Bi-LSTM outperforms the standard LSTM implementation on all lookback periods. We also consider a Random Forest baseline which upon inspection is found to generate sub-par learning capability on the failure data thereby reinforcing the need for learning temporal sequence patterns as our Bi-LSTM model does. At an accuracy of 96.4\% for a 15 day look-back, our model is among the state-of-the-art reported in the literature. An overview of our approach is shown in Fig. \ref{fig:workflow}.

%
%


\section{Previous Approaches}
In this section, we review some of the previous works focused on remaining useful life prediction of HDDs. For ease of review, we have divided these into statistical and machine learning approaches.

\subsection{\textbf{Statistical Approaches}}

Remaining useful life prediction is inherently a time series problem. The S.M.A.R.T. features of each hard drive are recorded at a constant time interval. Each S.M.A.R.T. attribute value is dependant on the previous value. This lack of independence makes this problem difficult to solve with simple statistical models such as linear regression. This section will briefly cover a few statistical approaches historically taken. For more details, the reader is encouraged to see the cited papers. 

One such approach (Wang et al.) involved using a two-step parametric method. The first step was transforming the variables into Gaussian variables with the Box-Cox transformation, using Mahalanobis distance to get the variables into one index. This was used for anomaly detection. Second, a generalized likelihood ratio test with a sliding window was used to predict failure. This method resulted in around a 68\% failure detection rate \cite{wang_ma_chow_tsui_2014}.

Another approach (Wang et al.) used an adaptive Rao-Blackwellized particle filter error tracking method. This gave a health status to the HDD. The failure was predicted by using a threshold that is placed by the adaptive error tracking. Using an accelerated degradation test, they were able to achieve a 97.44\% failure detection rate \cite{wang_he_jiang_chow_2021}. It should be noted that this approach was used to predict whether a hard disk drive would fail, not the RUL of an HDD.

A third approach (He, Yang and Xie) proposes using the Weibull distribution instead of the exponential distribution to predict RUL. Using sectional Weibull modeling can better capture the nuances of the HDD time-to-failure distribution \cite{6407546}. 

The above approaches rely heavily on a deep understanding of specific statistical techniques to predict impending failure. Model-driven approaches are highly pointed and they do not always end up learning the complex, underlying patterns in unstructured failure data with class imbalance.

\subsection{\textbf{Data-driven Approaches}}

In recent years, there has been an exponential rise in the adoption of data-driven approaches for fault prediction tasks. The solution frameworks involving the (a) prediction of RUL of an HDD instance and (b) classifying whether an HDD instance will fail within a certain number of days have both seen significant improvements with the adoption of machine learning and and deep learning approaches. This section will briefly survey some machine learning methods widely used in HDD RUL prediction.

A Bayesian Network is "a probabilistic graphical model that represents a set of random variables and their conditional dependencies" \cite{chaves_de}. These use the S.M.A.R.T. attributes to predict the probability that a hard drive will fail at a particular time. Simple (Chaves et al.) \cite{chaves_de, chaves_de_2016} and Combined Bayesian Networks (Jia et al.) \cite{pang_jia_stones_wang_liu_2016} have been used to predict the RUL. The Combined Bayesian Network is able to combine multiple machine learning classifiers to give a model that can predict more accurately than an individual classifier.

Relatively-straightforward machine learning models have also had some success in solving this problem. Machine learning algorithms such as Support Vector Machines (SVM), Decision Trees (DT), and Random Forests (RF) are frequently used. These algorithms are often used as a baseline or control to compare a proposed method to \cite{hu_han_xu_jiang_qi_2020}, as they are known to be well-performing. However, RF algorithms have shown to predict the RUL very well, and are commonly treated as more than a baseline \cite{shen_wan_lim_yu_2018, anantharaman_qiao_jadav_2018, aussel_jaulin_gandon_petetin_fazli_chabridon_2017}. Random Forest algorithms do not require a sequence of time data, so the input of current S.M.A.R.T. attribute values would be enough to predict the RUL \cite{anantharaman_qiao_jadav_2018}. This means that RF approaches need less data, offering an advantage over some of the time-dependant models discussed later. 

By far, the most common method to predict the RUL of hard drives involves using Long Short-Term Memory networks (LSTMs). These are an improvement on Recurrent Neural Networks that can help solve the vanishing gradient problem \cite{LSTM}. A detailed explanation of the LSTM and one of its variants is provided in the next section. LSTMs \cite{aussel_jaulin_gandon_petetin_fazli_chabridon_2017, basak_sengupta_dubey_2019, shen_ren_wan_lan_2021}, Convolutional Neural Network LSTMs \cite{246172}, Clustered LSTMs \cite{basak_sengupta_wen_dubey_2021}, and Attention LSTMs \cite{chen_wu_zhao_guretno_yan_li_2021, wang_wang_sun_2021} have all been used to predict hard drive failure or the RUL of a hard drive with a high degree of success.

A convolutional neural network LSTM (CNN-LSTM) is able to combine the benefits of both CNNs and LSTMs into one model. The CNN uses convolutional layers to reduce variation in the input. This filtered input is then passed to the LSTM, giving the LSTM better input than without the CNN. This combined model could give better results than a simple CNN or LSTM would alone \cite{246172} \cite{sainath_vinyals_senior_sak_2015}.

An encoder decoder LSTM, also known as a sequence-to-sequence model, uses two models. The first model encodes the input sequence to a fixed length. The second model decodes that fixed length input and predicts a sequence of output \cite{cho_van}. Remaining useful life prediction does not necessarily call for sequence prediction. All we need to predict is one number, the RUL. However, encoder decoder models have shown great performance and are often the base to an attention LSTM, so it is worth mentioning.

\section{The Long Short-Term Memory Network}

Recurrent neural networks (RNNs) are a form of neural networks that are commonly used for machine learning tasks involving a sequence of inputs. This could be anything from text classification \cite{rnn_ex1} to time series forecasting \cite{rnn_ex2}. However, recurrent neural networks suffer from the vanishing gradient problem which occurs when the network's weights are being updated based on the gradient during backpropagation. The gradient gets increasingly small as the training continues, eventually keeping the network from learning. To get around this problem, the Long Short-Term Memory Network (LSTM) was introduced \cite{LSTM}. 

\subsection{\textbf{Vanilla LSTM}}

\begin{figure}
    \centering
    \includegraphics[width=\columnwidth]{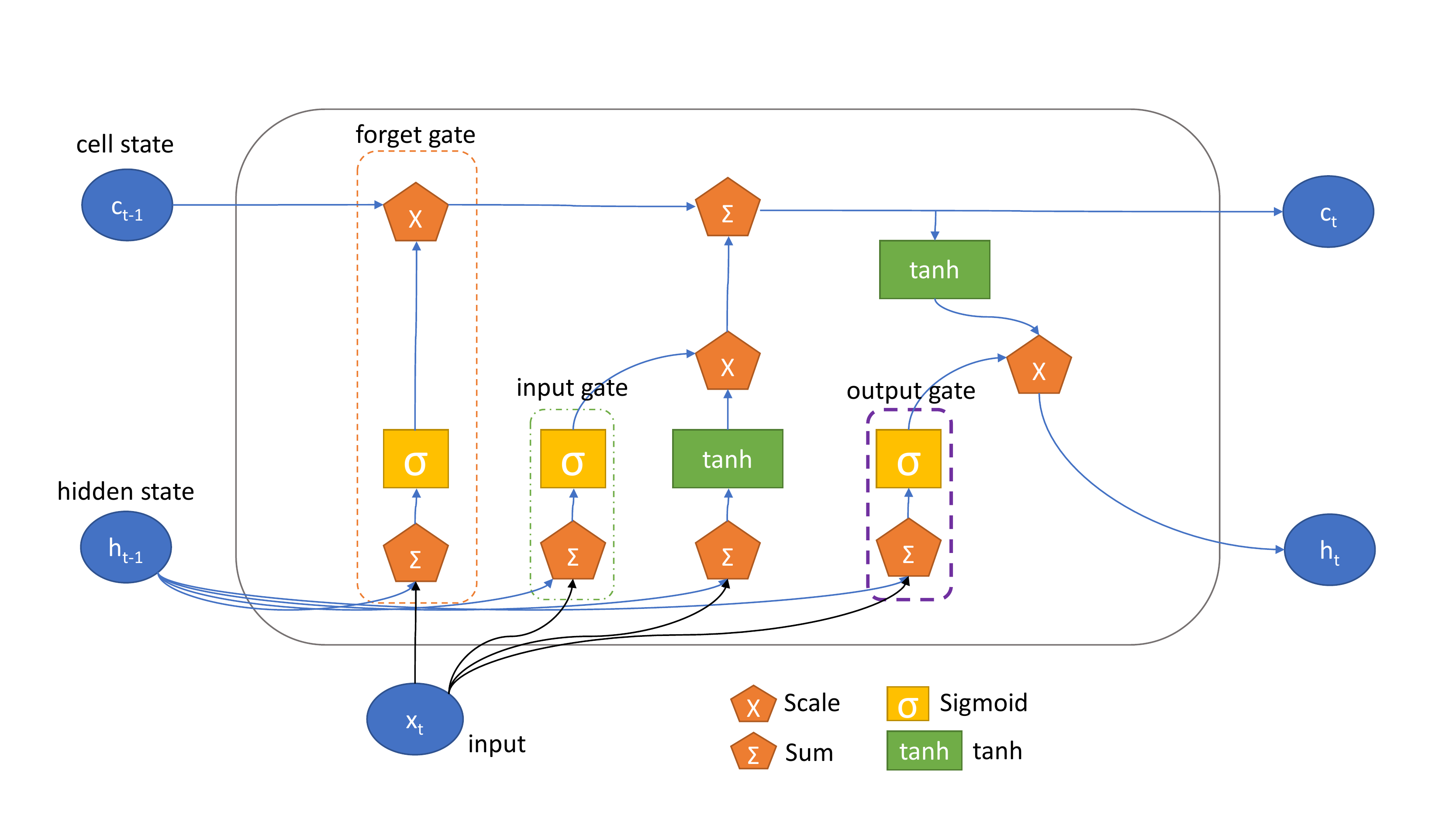}
    \caption{The Structure of an LSTM Cell}
    \label{fig:LSTM}
\end{figure}

The LSTM introduces a cell state that contains a series of gates in order to gain more control about the information that is retained between cells. The LSTM has a forget, input and output gate. The combination of these essentially allows the LSTM cells to extend its short term memory, keeping any information it needs to through the entirety of the learning process. Each gate contains neural networks that serve a specific purpose and contain activation functions such as sigmoid. The forget gate takes in information from the previous cell and current input to decide what to keep or forget. Whatever information is kept goes through the input gate. This determines what values will be updated in the cell. The $tanh$ function is applied on the cell state and current input for regulation. The cell state is then updated according to the combination of forget and input gates. Using the current cell gates and state, the output gate decides what to pass on to the next cell. A diagram outlining the LSTM cell is shown in Fig. \ref{fig:LSTM}.

\subsection{\textbf{Bidirectional LSTM}}

Numerous variants of LSTMs have been introduced to improve performance. We use one such variant - the Bidirectional LSTM or Bi-LSTM - in combination with specialized normalization strategies to predict and improve upon the HDD RUL as predicted by vanilla LSTMs.

\begin{figure}
    \centering
    \includegraphics[width=\columnwidth]{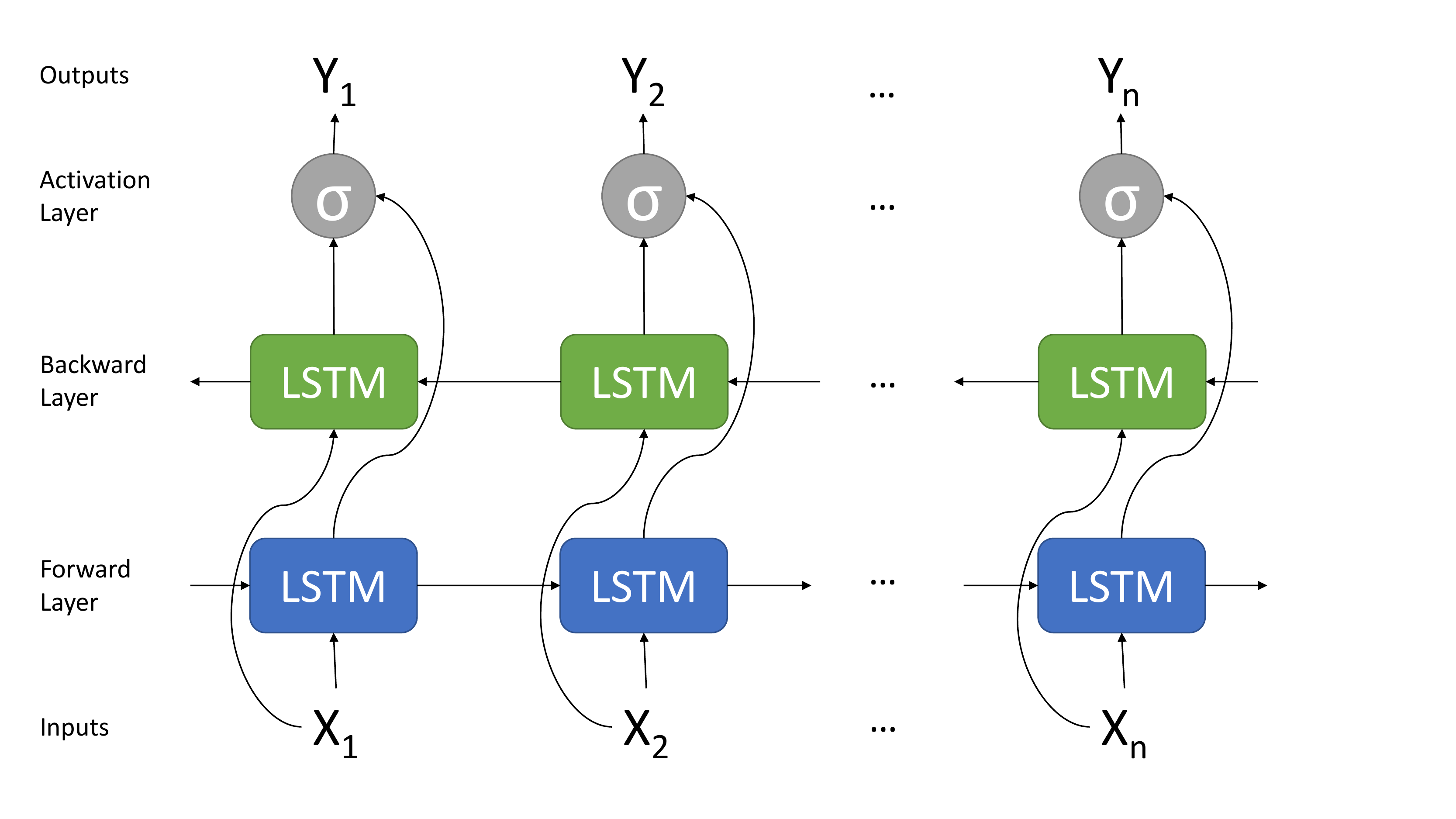}
    \caption{Bidirectional LSTM Architecture}
    \label{fig:BiDirectional}
\end{figure}

A bidirectional LSTM is a variant of an LSTM that consists of two LSTMs. These LSTMs run at the same time. One runs on the input sequence and the other runs backwards on the input sequence \cite{guo_wang_chen_ge_chen_2018}. In this way, the LSTM runs in both directions. In this problem, one could think of one direction of the LSTM running on the sequence of hard drive data leading up to failure and another running on the sequence as the hard drive gets further away from failure. This allows the LSTM to better learn the relationship between the features and the remaining useful life. Fig. \ref{fig:BiDirectional} shows the architecture of a bidirectional LSTM.

\section{Data Preprocessing} \label{The Data}

Backblaze is a cloud storage company that has over 170,000 HDDs used by customers across the globe. Each quarter, they release snapshots of all of their hard disk drives \cite{backblaze}. In these snapshots, there are daily statistics for each HDD including S.M.A.R.T. features and a few other features for each HDD. Among those is whether or not a HDD failed on a given day. If it failed, it it marked with a 1 and removed from the subsequent snapshots.

\subsection{\textbf{Data Collection}}
Using the Backblaze data from quarter 4 for 2019 and quarters 1 and 2 for 2020, we programmatically created a dataset containing the S.M.A.R.T. features leading up to failure for one hard drive model. Only one model was selected due to the inconsistency in meanings of S.M.A.R.T. features between models. The model selected was the Seagate ST4000DM000. We target this model because of the high prevalence of disk failures compared to other models.

This selection also allows for comparison with several existing state-of-the-art approaches in the literature such as Basak et al. \cite{basak_sengupta_wen_dubey_2021} and Anantharaman et al. \cite{anantharaman_qiao_jadav_2018}. Basak et al. used a spatio-temporal approach to predict the RUL of HDDs combining LSTM and hierarchical clustering. They achieved a mean absolute error (MAE) of 2.4 days for the prediction on Seagate ST4000DM000. Anantharaman et al. used two approaches: (a) a Random Forest model using the current snapshot of S.M.A.R.T. readings and (b) an LSTM which models the historical temporal pattern of the S.M.A.R.T. features. Their results suggest the Random Forest predictions are comparable to or outperforms the LSTM on Seagate ST4000DM000.

In order to create our dataset, we first looked for a hard drive with the model ST4000DM000 that had failed. When one was found, the previous 60 and 120 days worth of data were gathered for that same serial number. This was then concatenated onto the dataset in the long format. For each day back, a column for the remaining useful life was also added. For example, if a failed device was found, the column for RUL would have a value of 1 on the previous day, 2 on the day before that, and so on. This allowed us to keep from having to treat this as a binary classification problem. We could instead predict the RUL value using regression.

We chose to take a regression approach to this problem because we felt it remained true to the goal of the model. With regression, the model will be able to output a continuous number representing the predicted amount of days until the hard drive fails. Since the data is captured each day, this predicted number could be rounded or truncated. This will lead to a simple way to assess the performance of the model. One approach that was considered for binary classification was to iterate over each day, essentially asking the model the question "Will the hard drive fail on this day?" One could then use the first day where the model gives a value of 1 (hard drive failing) as the remaining useful life, capturing classification statistics along the way. We suspect this is how many previous approaches to RUL prediction handled this dataset, as most statistics provided in the literature are classification statistics (F Score, Precision, Recall, etc.).

\subsection{\textbf{Feature Selection}}
The Backblaze dataset provides dozens of S.M.A.R.T. statistics for each of its hard drives; however, not all of these are relevant. Some of the features are redundant, do not contain useful information, or are null. To remove these and reduce the complexity of the model, we used two feature selection methods. These were a correlation score and a Decision Tree.

Before using any feature selection techniques, we removed all null features and features that were already normalized. Backblaze performs their own data normalization and provides raw and normalized versions of each feature reported. We chose to apply our own normalization techniques, discussed later, and therefore removed all of these features.

\begin{figure}
    \includegraphics[width=\columnwidth]{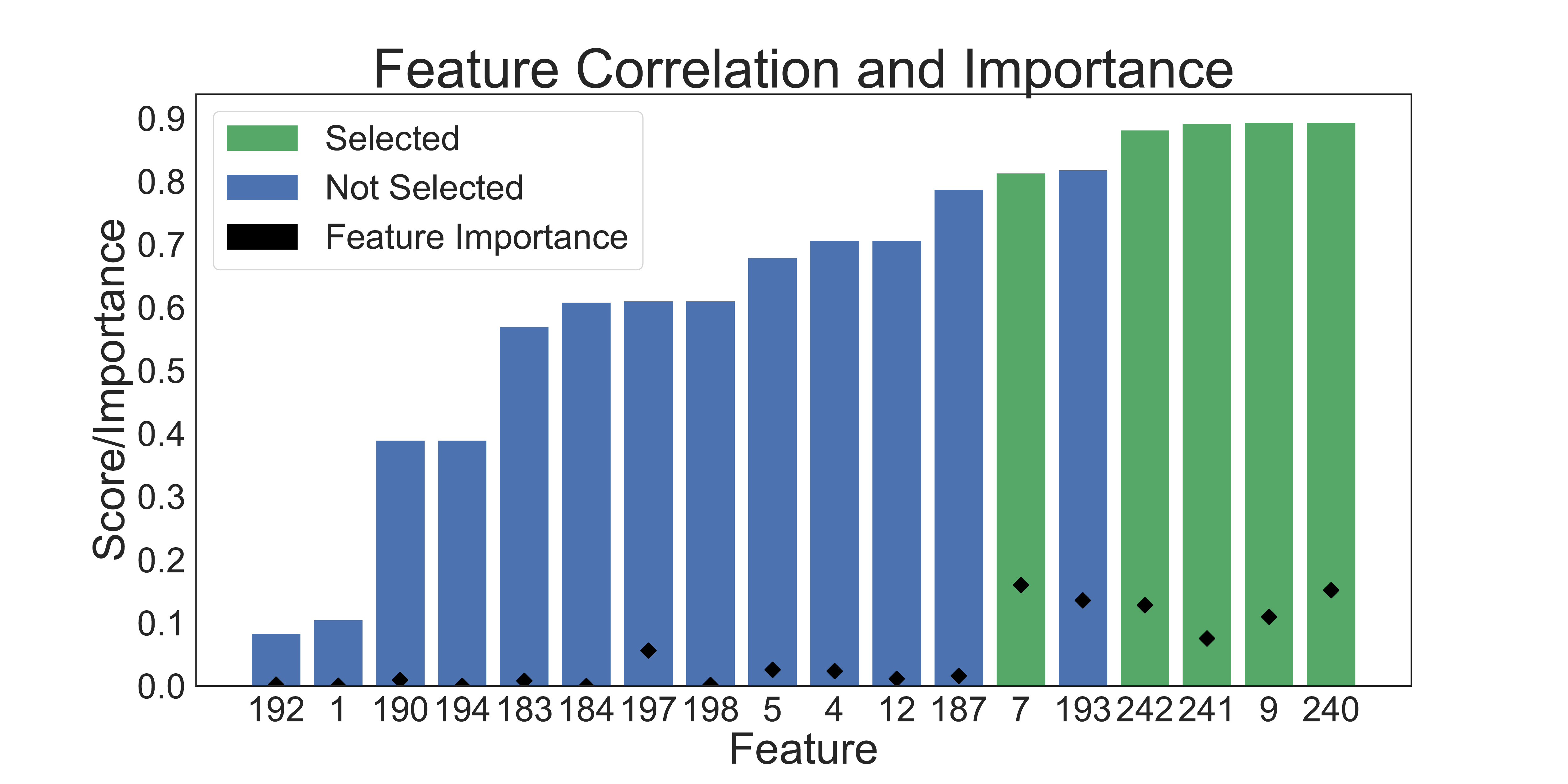}
    \caption{Pearson Correlation Score for S.M.A.R.T. Features as bars, Feature importance from Decision Tree as diamonds.}
    \label{fig:correlation}
\end{figure}

The first method of feature selection we chose was using a correlation score. We calculated the Pearson product-moment correlation coefficient between each S.M.A.R.T. feature and the feature representing the remaining life of the hard drive. This was calculated for each hard drive instance, averaged, and then the absolute value was taken. This left us with a correlation score for each S.M.A.R.T. feature, shown in Fig. \ref{fig:correlation}. As can be seen, 5 of the features are highlighted in green. These indicate the features that we selected as the final predictors. S.M.A.R.T. 7 was included over S.M.A.R.T. 193 because it was used as a predictor in \cite{basak_sengupta_wen_dubey_2021} and we wanted to make sure a fair comparison was possible. We suspect that 193 had a higher correlation score in our analysis than S.M.A.R.T. 7 due to differences in data, but the scores of S.M.A.R.T. 7 and S.M.A.R.T. 193 were practically the same. 

With concerns about nonlinearity of S.M.A.R.T. feature values, we conducted another method of feature selection. We trained a simple Decision Tree regressor on the dataset using all of the features. A graph demonstrating the importance it gave to each feature is shown on top of the bar plots in Fig. \ref{fig:correlation}. The diamonds represent the importance percentage of the respective feature determined by the decision tree. The sum of each of these importances is equal to one. As can be seen, the decision tree feature importance agrees with the correlation scores. S.M.A.R.T. 7, 240, 193, 242, 9, and 241 are given the most importance and highest correlation scores. Because of this, we felt comfortable taking S.M.A.R.T. 7, 9, 240, 241, and 242 as the features for the LSTM models. As discussed, these are the same features reported in \cite{basak_sengupta_wen_dubey_2021}, which will allow for more comparability in model performance.

A brief description of what each selected feature represents \cite{smart} is provided below. \\

\subsubsection{S.M.A.R.T. 7}
\textbf{Seek Error Rate}. The rate of seek errors of the magnetic head.

\subsubsection{S.M.A.R.T. 9}
\textbf{Power-on Hours}. The total number of hours the hard drive has been powered on.

\subsubsection{S.M.A.R.T. 240}
\textbf{Head Flying Hours}. The amount of time a disk head has spent moving.

\subsubsection{S.M.A.R.T. 241}
\textbf{Total LBAs Written}. The total number of LBAs written by the hard drive. An LBA is a 512 byte section of memory.

\subsubsection{S.M.A.R.T. 242}
\textbf{Total LBAs Read}. The total number of LBAs read by the hard drive. \\

It seems to make intuitive sense that each of these features would be highly correlated with hard drive failure.

\subsection{\textbf{Data Standardization}}
\begin{figure}
    \includegraphics[width=\columnwidth]{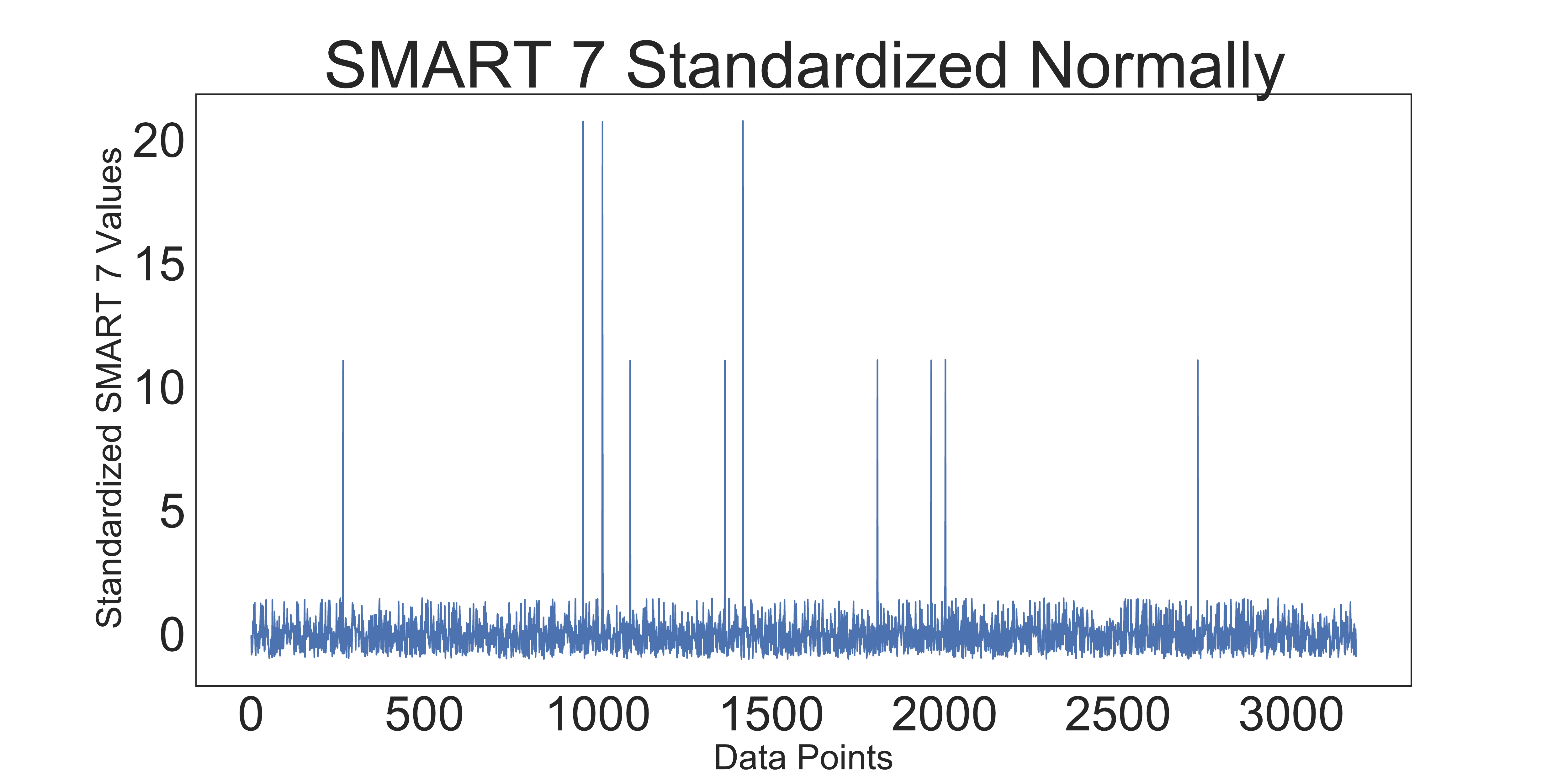}
    \caption{Values for S.M.A.R.T. 7 when standardized in a traditional way (Note the extreme points)}
    \label{fig:standardization}
\end{figure}

A typical method of standardization would involve scaling all features of all of the data points to some standard range. When done on this data, this led to a few points being more extreme than others, see Fig. \ref{fig:standardization}. As can be seen, a few of the points have values above 10. The standardization makes the points have a mean of 0 and variance of 1. All of the points should ideally be somewhere between -2 and 2 to have an even distribution. The points that are extremely far outside of this range may inhibit the LSTM's training and reduce performance. 

In order to fix this, a different approach to standardization was taken. Instead of standardizing each feature from every hard drive (the full dataset), each individual hard drive was standardized independently of the other. Each feature from each hard drive was scaled such that they would have a mean of 0 and variance of 1. This heavily reduced the impact of the extreme data points. However, this did come with a cost. We could no longer apply this same scaler to the test set since each scale was dependent on the hard drive. This has potential impacts on 'across-the-board' generalization of the model, a topic which will be discussed further in the paper.

Since applying the same scaler to the training and test sets was no longer an option, we took the same approach to scaling the test set and other validation sets as the training set. Each feature for each device was independently scaled to a mean of 0 and variance of 1. 

This approach can be applied to any hard drive data of the same shape. This means that someone using this model could easily standardize their data and run it through our model without having to apply the exact transformation as us, something one would most likely have to do with typical normalization.

An alternative solution to this issue, and one that has been applied in the past \cite{basak_sengupta_wen_dubey_2021}, would have been to perform some kind of outlier detection and removal. However, this comes with the cost of losing potentially valuable data. Some points labled as outliers would be easy to justify as impossible. For example, consider a hard drive that reports its hours active as longer than the hard drive has been manufactured. Other points would not be as easy to justify removing, and there is the potential of removing something useful to the model.

It also raises the question of how this would be handled in an actual implementation of the model. Would an end user have to be okay with a model simply being wrong about their hard drive's RUL if it reported one of these extreme points? If a model is trained on a dataset without these extreme points, we cannot expect it to be accurate when it inevitably encounters feature values like this outside of training. For these reasons, we did not use this approach.

\subsection{\textbf{Reclassification of Data}}
At this point, we had a set of training data consisting of 78 hard drives with 5 S.M.A.R.T. features reported for each hard drive 60 days up to failure and the day of failure. We had 2 sets of test data. One consisted of 71 hard drives, different than the ones in the training set, with the same 5 S.M.A.R.T. features reported for each drive 60 days up to failure and the day of failure. The second test set consisted of 133 hard drives with the same 5 S.M.A.R.T. features reported 120 days up to failure and the day of failure. 

\begin{figure}
    \includegraphics[width=\columnwidth]{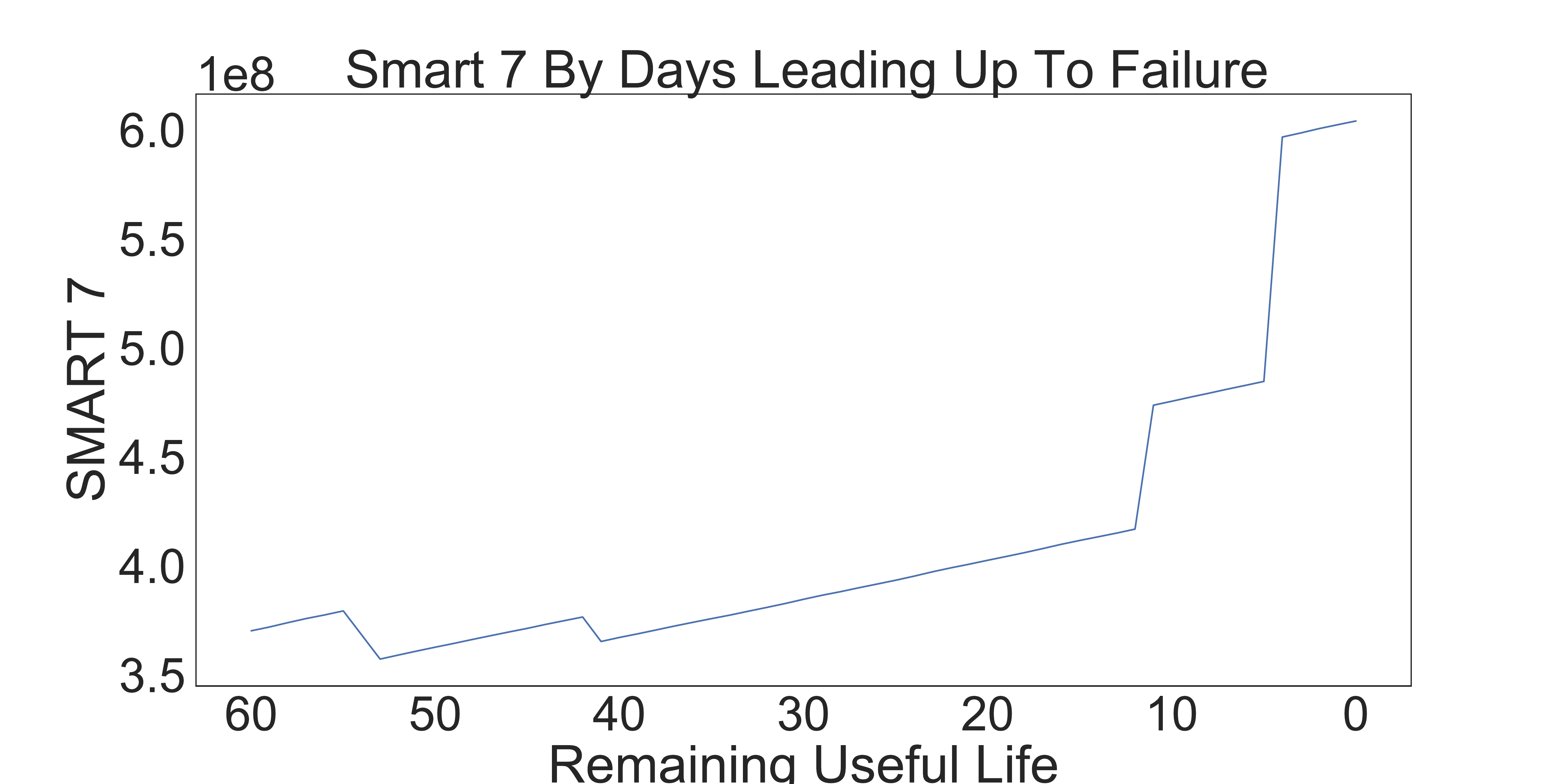}
    \caption{S.M.A.R.T. 7 values 60 days before failure (Notice the jump in values occurring about 15 days before failure)}
    \label{fig:smart7}
\end{figure}

After visualizing the data for each day leading up to failure, a pattern begins to emerge (see Fig. \ref{fig:smart7}). It appears that a few of the reported S.M.A.R.T. features have large jumps leading up to the day of failure. This is what would be expected. For example, consider S.M.A.R.T. 7, the rate of seek errors. One would expect these errors to increase dramatically as the hard drive begins to fail. Our model will likely be predicting a lower remaining useful life of the hard drive as each feature has a dramatic change. This will be a problem when trying to predict the RUL accurately using regression. How could the model say that a hard drive has 117 days of RUL instead of 116 when the feature values for these are virtually the same and these dramatic changes do not happen until right before failure?

To solve this problem, we reclassified every value for the remaining useful life that is over 30 days as 30 days. This created a category where the hard drive is in good working condition. This reclassification of the RUL is also done in \cite{anantharaman_qiao_jadav_2018}. Any day where a hard drive is reporting normal S.M.A.R.T. feature values can be treated as having 30+ days of remaining useful life. We believe this is appropriate given the inspection of the feature values and would still maintain the practicality of the model. A consumer using this model would have about 30 days to back up and replace their hard drive before failure, much better than the last minute notification of failure usually given.

\section{Model Training}

\subsection{\textbf{LSTM Models}}
To train an LSTM model, the data has to be in the shape of [samples x timesteps x features]. The samples are the number of data points. (Given 60 days worth of data, including a 0th day of failure, for 78 hard drives, the number of samples is 4758.) The timesteps are the number of days of data leading up to each sample point. We trained these LSTM models with a variety of parameters for the timesteps. With these timesteps, our models do not have to make their prediction based on the current S.M.A.R.T. feature values. The models can also use the previous $n$ days' S.M.A.R.T. features to better understand the way the data is changing. The final dimension of the shape that LSTM data has to be in is the number of features. In this case, this is simply 5, as we chose 5 S.M.A.R.T. features in the previous section.

With the data in the correct shape, the LSTM models could be trained. To train the basic LSTM, we created a model with an LSTM layer consisting of 32 units followed by a densely-connected neural network layer. The Adam optimizer was chosen. The model was trained for 50 epochs with a batch size of 64 and was shuffled during training. A very similar approach was taken to train the bidirectional LSTM. A Bidirectional wrapper was added to the LSTM layer that would return the last output.

\begin{figure}
    \includegraphics[width=\columnwidth]{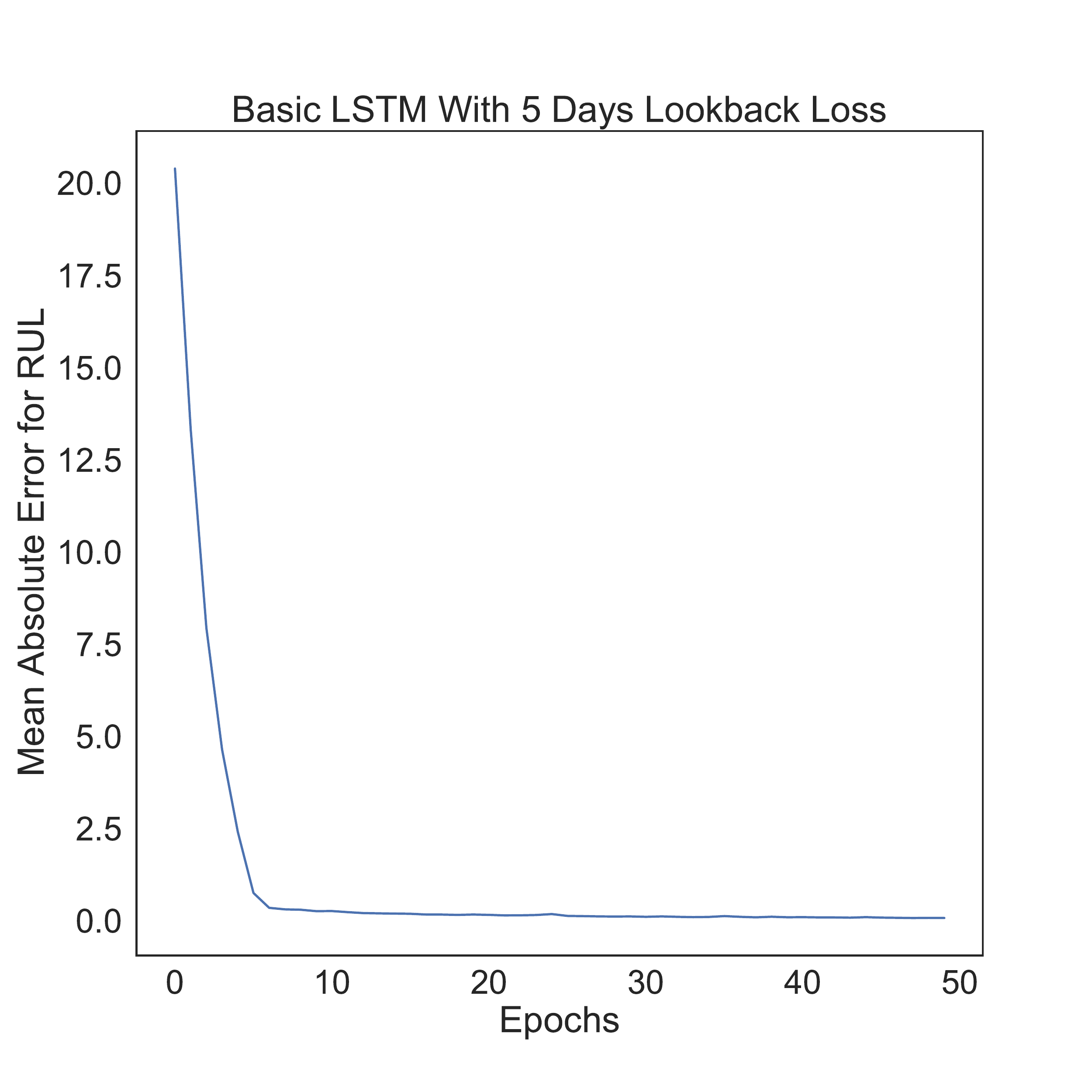}
    \caption{Loss function for a vanilla LSTM trained with 5 days look back over 50 epochs. Loss curves of other LSTM models follow a similar trend.}
    \label{fig:loss}
\end{figure}

All of the LSTM models were trained with 5, 10, 15, and 30 timesteps. This was done to see the impact on RUL prediction when the model is given varying amounts of data. An example loss function graph for the LSTM training is provided in Fig. \ref{fig:loss}.

\subsection{\textbf{Random Forest}}
To provide a baseline performance comparison, a Random Forest model was also trained. The random forest was trained on the same dataset as the LSTM models, split into 80\% training and 20\% testing. The data for the random forest was not standardized and each feature was given to it. Due to the flexibility of random forests, this would have been unnecessary and led to reduced performance when we tried it. Any data point more than 30 days away from failure was still capped at 30 days for comparability. It was trained with 1000 estimators and no max depth. 

\section{Results}
After each model was trained, each model was then evaluated on two test sets. The first included data similar to the data the models were trained on. This test set had the 5 S.M.A.R.T. feature values for the 60 days leading up to a hard drive failure for 71 hard drives. The second test set was included to see how to the trained models would perform when given a more difficult task. Instead of predicting the RUL for data at most 60 days away from hard drive failure, the second test set included data 120 days leading up to each hard drive's failure. This dataset included 133 hard drives. Both of these test sets were standardized in the same way the training sets were.

\begin{table}[htbp]
    \centering
    \caption{Performance for trained models on test data 60 days before failure.}
    \begin{tabular}{@{}lllll@{}}
        \toprule
        Model              & Timesteps & Accuracy & R$^2$ & MAE   \\ \midrule
        LSTM         & 5         & 0.910    & 0.992                & 0.287 \\
        LSTM         & 10        & 0.914    & 0.994                & 0.244 \\
        LSTM         & 15        & 0.916    & 0.992                & 0.247 \\
        Bi-LSTM         & 30        & 0.910    & 0.994                & 0.238 \\
        Bi-LSTM & 5         & 0.939    & 0.999                & 0.131 \\
        Bi-LSTM & 10        & 0.934    & 0.997                & 0.190 \\
        \textbf{Bi-LSTM} & \textbf{15}        & \textbf{0.964}    & \textbf{0.998}                & \textbf{0.120} \\
        Bi-LSTM & 30        & 0.960    & 0.998                & 0.132 \\
        RF                 & NA        & 0.667    & 0.987                & 0.483 \\ \bottomrule \\
    \end{tabular}
    \label{tab:60days}
\end{table}

In order to evaluate the performance of these models, we needed some quantitative metrics. Because our models would output a continuous number representing the predicted remaining useful life, we chose R$^2$ as one of these metrics. We also collected the mean absolute error (MAE) in the predicted RUL. To concretely represent the usefulness of the models, we also collected accuracy. To determine how many correct predictions the model made, we rounded each prediction. So a prediction of 24.4 days RUL when the hard drive actually had 24 would be deemed a correct prediction.

The performance of each LSTM model with 5, 10, 15, and 30 days timesteps for 60 days of data is shown in TABLE \ref{tab:60days}. The basic LSTM network is denoted as LSTM and the bidirectional LSTM is denoted as Bi-LSTM.


The performance of each LSTM model with the same timesteps for 120 of data is shown in TABLE \ref{tab:120days}. The random forest model is also added for comparison at the bottom of both tables.



As can be seen, the LSTM models achieve high accuracy on data within the same range it is trained on. In TABLE \ref{tab:60days} we see that the Bidirectional LSTM with a timestep of 15 days performs the best on the test set. This model achieves a MAE of 0.12, much lower than most of the state-of-the-art models, including Basak et al. \cite{basak_sengupta_wen_dubey_2021} that achieved a MAE of 2.4 on the same model. Every LSTM model outperforms the Random Forest model. All of the LSTM models are able to generalize very well to the same time frame.

\begin{table}[htbp]
    \centering
    \caption{Performance for trained models when extrapolating to 120 days away from failure.}
    \begin{tabular}{@{}lllll@{}}
        \toprule
        Model              & Timesteps & Accuracy & R$^2$ & MAE   \\ \midrule
        LSTM         & 5         & 0.479    & 0.286                & 4.305 \\
        LSTM         & 10        & 0.471    & 0.219                & 4.589 \\
        LSTM         & 15        & 0.444    & 0.028                & 5.250 \\
        LSTM         & 30        & 0.358    & -0.359               & 6.433 \\
        Bi-LSTM & 5         & \textbf{0.497}    & 0.312                & 4.145 \\
        Bi-LSTM & 10        & 0.496    & 0.216                & 4.448 \\
        Bi-LSTM & 15        & 0.487    & 0.071                & 4.874 \\
        Bi-LSTM & 30        & 0.369    & -0.565               & 6.792 \\
        RF                 & NA        & 0.378    & \textbf{0.333}                & \textbf{3.647} \\ \bottomrule \\
    \end{tabular}
    \label{tab:120days}
\end{table}

Fig. \ref{fig:lstm_predictions}, Fig. \ref{fig:lstm_15_predictions}, and Fig. \ref{fig:rf_predictions} demonstrate the difference in performance between the LSTM and random forest models. All of these graphs are sorted by the actual remaining useful life for demonstrative purposes.

TABLE \ref{tab:literature_performance} shows performance metrics for related data-driven approaches for hard drive health assessment. While the table is not exhaustive, it should provide some context to the performance of state-of-the-art models.

\begin{table}[]
    \centering
    \caption{Performance of previous approaches for RUL prediction by various data-driven approaches.}
    \label{tab:literature_performance}
    \begin{tabular}{@{}lllll@{}}
    \toprule
    Model Type     & Precision & Recall & MAE  & Citation               \\ \midrule
    Random Forest  & 0.95      & 0.67   &  - & Aussel et al.\cite{aussel_jaulin_gandon_petetin_fazli_chabridon_2017}          \\
    Attention LSTM & 0.93      & 0.96   &   -   & Wang et al.\cite{wang_wang_sun_2021} \\
    Random Forest  & 0.66      & 0.94   &   -  & Lu et al.\cite{246172}               \\
    LSTM           & 0.66      & 0.88   &   -   & Lu et al.\cite{246172}               \\
    CNN-LSTM       & 0.93      & 0.94   &   -  & Lu et al.\cite{246172}               \\
    Random Forest  &    -       &  -      & 6.4  & Anantharaman et al.\cite{anantharaman_qiao_jadav_2018} \\
    LSTM           &    -       & -       & 8.15 & Anantharaman et al.\cite{anantharaman_qiao_jadav_2018} \\
    Clustered LSTM &    -       &  -      & 2.4  & Basak et al.\cite{basak_sengupta_wen_dubey_2021} \\
    Bi-LSTM &    -       &  -      & 0.12  & \textbf{Our Approach}\\
    \bottomrule
    \end{tabular}
\end{table}

\subsection{\textbf{Observations}}

\begin{figure}
    \includegraphics[width=\columnwidth]{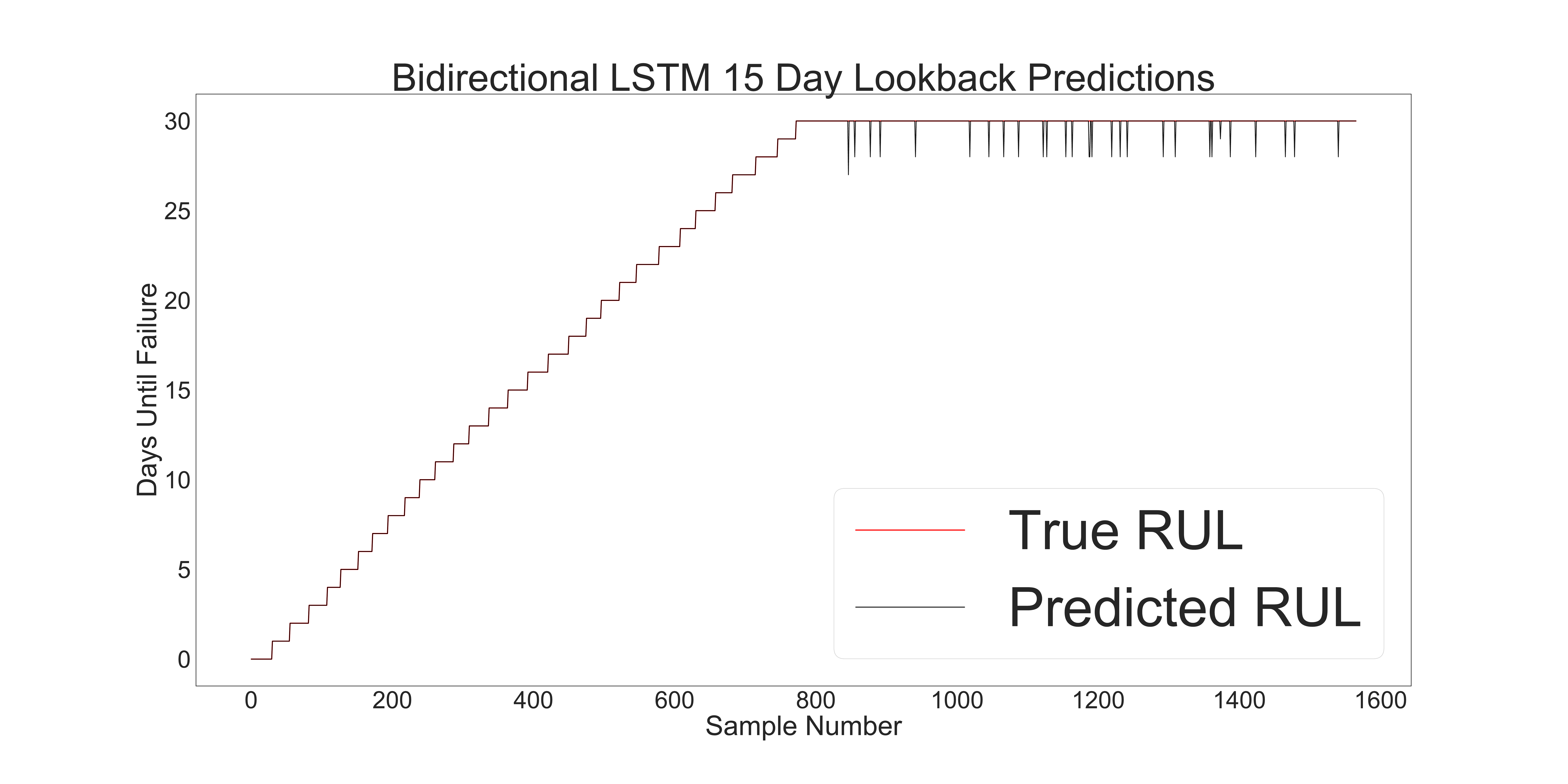}
    \caption{Bidirectional LSTM with 15 days lookback predictions. The points are sorted by the actual RUL.}
    \label{fig:lstm_predictions}
    
    \includegraphics[width=\columnwidth]{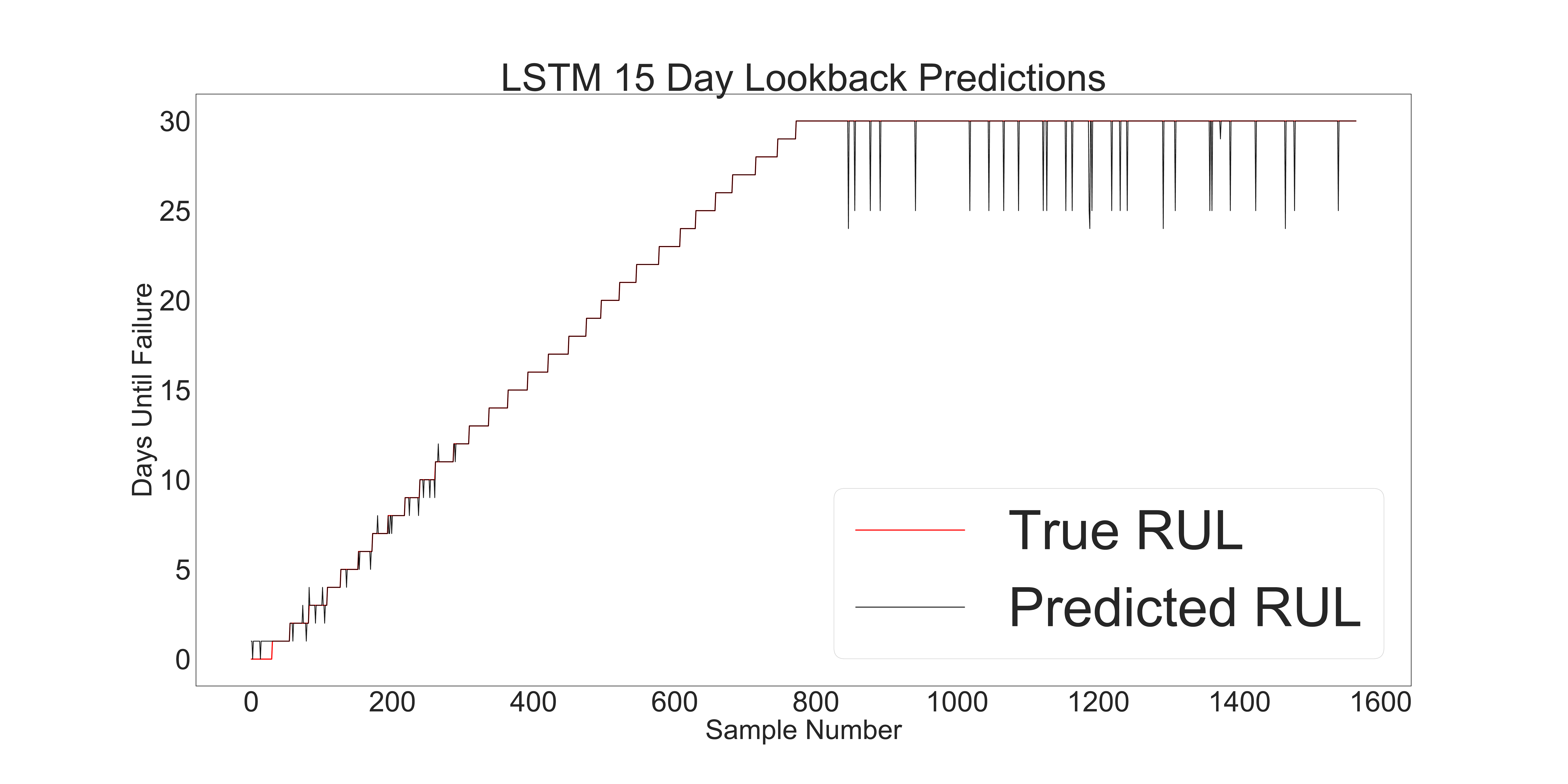}
    \caption{LSTM with 15 days lookback predictions. The points are sorted by the actual RUL.}
    \label{fig:lstm_15_predictions}
    
    \includegraphics[width=\columnwidth]{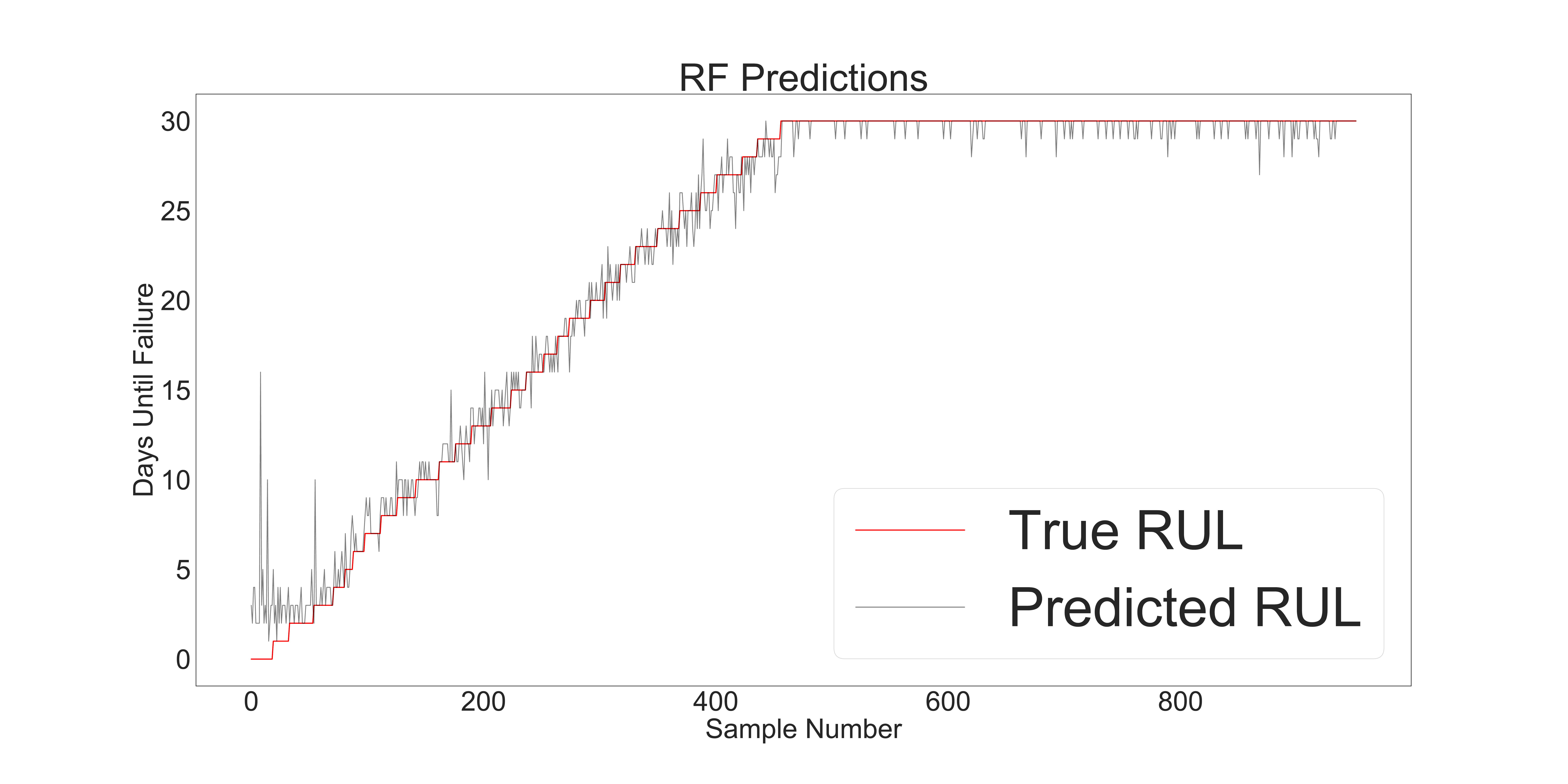}
    \caption{Random Forest predictions. The points are sorted by the actual RUL.}
    \label{fig:rf_predictions}
\end{figure}

Some interesting observations can be gained from the graphs in Fig. \ref{fig:lstm_predictions}, Fig. \ref{fig:lstm_15_predictions}, and Fig. \ref{fig:rf_predictions}. First, it is clear that the LSTM models outperform the Random Forest. While Fig. \ref{fig:lstm_predictions} shows almost no separation between the True RUL and the Predicted RUL for the Bi-LSTM architecture, the difference is most clearly visible in Fig. \ref{fig:rf_predictions} for the Random Forest. It is also clear that the Bi-LSTM outperforms the regular LSTM model in Fig. \ref{fig:lstm_15_predictions}. Second, we see that the LSTM models incorrectly predict the RUL more often when the hard drive is further away from failure. Most of the deviations from the actual remaining useful life are when the hard drive is in the category of 30+ days RUL. The LSTMs are more accurate when the hard drive is closer to failure, reinforcing its practical importance. Third, we can see that the Random Forest appears to have an opposite relationship with the data. The Random Forest is much less accurate the closer the hard drive gets to the day of failure. There is not a spot on Fig. \ref{fig:rf_predictions} where the Random Forest appears to consistently predict the RUL, but it is certainly worse when the actual RUL gets below 10 days. It is also worth noting that the Random Forest is appearing to overestimate the remaining useful life. A real user would probably want the model to underestimate the RUL. It is better that they replace their hard drive early than have it fail on them before they can replace it.

Every model performs poorly when attempting to extrapolate far into the future. This is a common problem with statistical and machine learning prediction methods in general. From TABLE \ref{tab:120days}, we see that the Bidirectional LSTM has the highest accuracy at predicting the remaining useful life of hard drives that are twice as far away from failing (120 days) as the models were trained on (60 days). This LSTM model gets nearly half of its predictions correct. In contrast, the random forest has a lower accuracy score, but also has a higher R$^2$ and lower MAE. This implies that the random forest predictions may actually be better than the LSTM models, but the random forest does not predict the exact RUL as well. 

It would be hard to expect a model to be able to perform as well on data that is twice as far away as the data it was trained on, but being able to do so would highly improve the real-world usefulness of the model.

\section{Standardization and Generalization}
This section is dedicated to more discussion of our novel standardization approach. The benefits and downsides of this method of standardization will be explained.

As mentioned in Section \ref{The Data} C, the hard drive data needed to be normalized or standardized to be used by the LSTM models. A traditional standardization approach where the entire features would be standardized as a whole was considered, but not adopted due to the poor LSTM performance. We believe the cause of this to be partially due to extreme data points influencing the weights of the LSTM and difficulty of the problem. 

To standardize the data, we standardized the data per sample. This means that each feature of each hard drive was essentially treated as its own dataset when applying the scaler. 

To better explain this, consider two hard drives: hard drive A and hard drive B. Both hard drive A and B are of the same model, but hard drive A has an extreme value for its S.M.A.R.T. 7 feature on 2 of the 60 days of data captured. When all of the S.M.A.R.T. 7 feature values are standardized together, the weight of these extreme values are maintained. The standardization we used made each feature have a mean of 0 and variance of 1. It follows that the more data points that exist, the less the extreme values of S.M.A.R.T. 7 that hard drive A contains will be pulled towards a reasonable number. For some models, such as a Random Forest, this may not be an issue, but because of the nature of recurrent neural networks these extreme values will amplify the weights and reduce the performance of the model.

\begin{figure}
    \centering
    \includegraphics[width=\columnwidth]{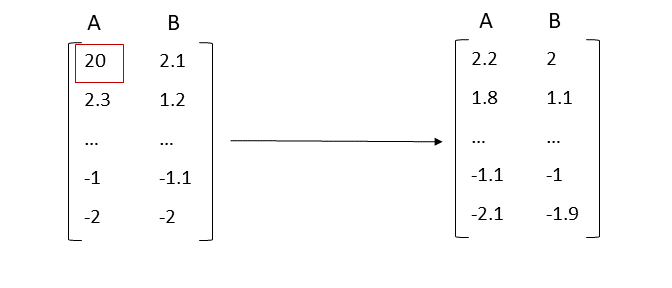}
    \caption{Visualizing a potential difference between the regular method of standardization (left) and the per sample standardization (right.)}
    \label{fig:matrix}
\end{figure}

By normalizing the S.M.A.R.T. 7 values for hard drive A and B separately, the extreme values for hard drive A will be pulled more towards reasonable values while still maintaining the patterns in their feature values that the LSTM can learn. Fig. \ref{fig:matrix} is included to visualize this process. The data included in this figure is entirely theoretical, but demonstrates the differences in the range of the data for both of the standardization techniques discussed. 

This method of standardization deals with the extreme data points without risking a loss of data as would be done using an outlier removal. However, this does have the potential downside of not being able to reuse the same standardization function to new data. At first, we assumed that this would cause the model to lose any generalization. From the results, we see that this was not the case. The LSTM models performed exceptionally well at predicting the RUL of hard drives after undergoing this standardization. It outperforms a model, the Random Forest, that was not standardized this way. 

As was also demonstrated, this method of standardization has the potential of restricting the range of time that the model is useful. We saw that the models did not perform well on 120 days worth of data. Using this standardization technique, the data around day 120 should have around the same values as the data around day 60 for the other test sets. Instead of being able to apply a standardization function that would potentially assign different values to those past 60 days, they are all standardized to a similar range. 

We believe that this should not discredit this technique of standardization. The performance of the LSTM models were greatly increased by this standardization technique over a traditional one. We attribute this to the complexity of the dataset. Many of the S.M.A.R.T. feature values remain relatively-unchanging until the hard drive is soon to fail. The typical method of standardization may assign a similar value for one of the S.M.A.R.T. features 60 days away from failure to one 120 days away from failure due to the nature of the data (see Fig. \ref{fig:smart7}). This, while still maintaining outliers, would probably make training an accurate LSTM very difficult. 

\section{Conclusion}

In this paper, we attempted to solve the problem of remaining useful life prediction for hard drives. Given a series of hard drive data 60 days before the hard drive fails, we were able to predict the remaining useful life of a hard drive with state-of-the-art accuracy, outperforming previous approaches in the literature \cite{basak_sengupta_wen_dubey_2021} that worked on the same data and the same model, in particular. 

To accomplish this, we used both Bidirectional and vanilla Long Short-Term Memory networks, or LSTMs, with highly customized pre-processing directives. We provided a cap on the number of days a hard drive could be away from failure, giving our model a category where the hard drive is treated as operating normally. This allowed the model to better handle data where the hard drive is not failing, improving the performance. We implemented a unique method of data standardization that standardized each hard drive independently. This maintained generalization on similar time windows, but came with the cost of reducing performance on time windows much different than trained on. 

We also implemented a Random Forest model as a baseline to compare our LSTM models to. The random forest model was able to perform as well as the LSTM models on data in a different time window than trained on, but was severely outperformed on the test data from the same time window. 

\subsection{\textbf{Future Work}}
While our work is able to predict the remaining useful life of hard drives well, there is still much room for future work to be done on this complex problem. Most notably, an ideal model would be able to achieve similar performance to our model while still being able to extrapolate well. There is no set number of days that one could select for training the model that would not run into extrapolation issues. Even if one were to train a model on 1,000 days of data, there is still a scenario where someone would have a hard drive that is more than 1,000 days away from failing. We attempted to solve this by implementing the cap on the remaining useful life, but the models still struggled to perform well when the time interval was doubled. Improving the extrapolation performance would be an important milestone towards true generalization and add to the applicability of the model in a real-world deployment scenario.

Second, future work could explore a model that is able to predict the remaining useful life for any hard drive model. While we believe our process could easily be reproduced for any hard drive model, it would be worth exploring the possibility of training a model that would work for more than one model. 

Finally, future work could implement a hard drive RUL prediction model into a real system, testing the performance of the model on real-time hard drive data for hard drives that may not fail for years.

\bibliographystyle{./bibliography/IEEEtran}
\bibliography{./bibliography/IEEEabrv,./bibliography/references}

\end{document}